\documentclass[11pt]{article}

\usepackage[preprint]{acl}

\usepackage{times}
\usepackage{latexsym}

\usepackage{minitoc}
\usepackage{titletoc}

\usepackage[T1]{fontenc}

\usepackage[utf8]{inputenc}

\usepackage{microtype}

\usepackage{booktabs,multirow,graphicx,array,makecell}

\newcolumntype{L}[1]{>{\raggedright\arraybackslash}p{#1}}
\newcolumntype{C}[1]{>{\centering\arraybackslash}p{#1}}
\newcolumntype{R}[1]{>{\raggedleft\arraybackslash}p{#1}}

\newcolumntype{Y}[2]{>{\hspace*{#2}\centering\arraybackslash}p{#1}}

\usepackage{graphicx}
\usepackage{amsfonts}
\usepackage{amsmath,amssymb}

\usepackage{multirow}

\usepackage{hyperref}
\definecolor{refcolor}{HTML}{9F363A}
\definecolor{custompurple}{RGB}{165,119,233}

\hypersetup{
    colorlinks=true,
    linkcolor=refcolor,
    citecolor=refcolor,
    filecolor=magenta,      
    urlcolor=refcolor,
    }
\usepackage{tikz}
\newcommand*\circled[1]{\tikz[baseline=(char.base)]{
            \node[shape=circle,draw,inner sep=0.4pt] (char) {#1};}}
\title{What Makes a Good Curriculum? Disentangling the Effects of \\ Data Ordering on LLM Mathematical Reasoning}

\author{
  \textbf{Yaning Jia, Chunhui Zhang, Xingjian Diao, Xiangchi Yuan}\\
  \textbf{Zhongyu Ouyang, Chiyu Ma, and Soroush Vosoughi}\\
  Dartmouth College \\
  \texttt{yaning.jia.gr@dartmouth.edu, soroush.vosoughi@dartmouth.edu}
}

\begin{document}
\maketitle
\begin{abstract}


Curriculum learning (CL)—ordering training data from easy to hard—has become a popular strategy for improving reasoning in large language models (LLMs). Yet prior work employs disparate difficulty metrics and training setups, leaving open fundamental questions: \textit{When does curriculum help? Which direction—forward or reverse—is better? And does the answer depend on what we measure?} We address these questions through a unified offline evaluation framework that decomposes curriculum difficulty into five complementary dimensions: \textit{Problem Difficulty}, \textit{Model Surprisal}, \textit{Confidence Margin}, \textit{Predictive Uncertainty}, and \textit{Decision Variability}. Through controlled post-training experiments on mathematical reasoning benchmarks with \textit{Llama3.1-8B}, \textit{Mistral-7B}, and \textit{Gemma3-4B}, we find that: \textit{(i)}~no curriculum strategy dominates universally—the relative effectiveness of forward versus reverse CL depends jointly on model capability and task complexity; \textit{(ii)}~even within a single metric, samples at different difficulty levels produce distinct gains depending on task demands; and \textit{(iii)}~\textit{Task-aligned} curricula focus on shaping the model’s final representations and generalization, whereas \textit{inner-state} curricula modulate internal states such as confidence and uncertainty. Our findings challenge the notion of a universal curriculum strategy and offer actionable guidance across model and task regimes, with some metrics indicating that prioritizing decision-uncertain samples can further enhance learning outcomes.
\end{abstract}

\section{Introduction}
Curriculum Learning (CL), which organizes training data by difficulty, has become an effective approach for improving the reasoning ability of large language models (LLMs)~\cite{bengio2009curriculum, kumar2010self, jiang2015self, gan2021self, feng2025your}. By presenting examples in a structured progression, CL has demonstrated improvements across diverse reasoning domains, including code generation~\cite{nair2024curriculum, pouransari2024dataset}, instruction following~\cite{lee-etal-2024-instruction, pang2024phased}, and commonsense reasoning~\cite{maharana2022curriculum}. Yet despite its widespread adoption, a fundamental question remains unresolved: \textit{what makes a curriculum strategy effective?}


The answer is far from straightforward. Existing work employs a heterogeneous landscape of difficulty metrics that fall into two broad families. \textit{Problem-side metrics} quantify intrinsic task complexity—such as reasoning depth~\cite{jung2025reasoning}, linguistic complexity~\cite{liu2008dependency}, symbolic density~\cite{polu2022formal, meadows-etal-2024-symbolic}, or model-independent success rates like Acc@\textit{K}~\cite{tong2024dart}. These metrics enable curricula that progress from structurally simple to complex problems, often improving both efficiency and accuracy~\cite{jung2025reasoning, ding2024easy2hard, parashar2025curriculum}. \textit{Model-side metrics}, in contrast, capture difficulty from the model's perspective—measuring uncertainty~\cite{zhou2020uncertainty,peng2023token}, confidence~\cite{pattnaik2024enhancing}, or surprisal~\cite{pattnaik2024enhancing, mohiuddin2022data}. By adaptively focusing on samples that challenge the model, these metrics can better expose knowledge gaps and enhance generalization in high-difficulty regimes.

However, prior work varies substantially in training protocols, evaluation setups, and even the direction of curriculum ordering (forward vs. reverse), making systematic comparison difficult. Many studies adopt online or reinforcement-learning–based curricula that dynamically adjust sampling~\cite{chen2025self, li2025curriculum, zhang2025speed}, confounding the effect of data ordering with optimization dynamics or reward shaping. Moreover, the choice of difficulty metric is often treated as a design decision rather than an empirical question, leaving practitioners without clear guidance on which metric to use, when forward or reverse ordering is preferable, or how these choices interact with model capability and task complexity.


We address these gaps through a controlled, multidimensional evaluation of curriculum strategies in mathematical reasoning. We decompose curriculum difficulty into five complementary metric families: \textit{Problem Difficulty}, \textit{Model Surprisal}, \textit{Confidence Margin}, \textit{Predictive Uncertainty}, and \textit{Decision Variability}, and systematically construct both forward (easy to hard) and reverse (hard to easy) curricula for each. Crucially, we focus on \textit{supervised fine-tuning (SFT)} under an \textit{offline} setting, where training data are fixed but reordered before learning. This design isolates the effect of data ordering by holding objectives, budgets, and optimization procedures constant, providing a clean lens to study curriculum effects without confounds from adaptive sampling or reward engineering.


We conduct controlled post-training experiments on standard mathematical reasoning benchmarks using three representative open-source base models—\textit{Llama-8B}, \textit{Mistral-7B}, and \textit{Gemma-4B}—spanning different scales and architectural families. This enables us to examine not only \textit{whether} curricula help, but \textit{when}, \textit{why}, and \textit{for whom} they are most effective.
Our findings reveal four key insights that challenge conventional lessons about curriculum learning:

\noindent \circled{1}~\textbf{Data ordering is a first-class learning signal.} Even when the training budget and data composition are fixed, curriculum structure alone can substantially alter learning dynamics and reasoning outcomes, highlighting that \textit{how} we present data matters as much as \textit{what} we present.

\noindent \circled{2}~\textbf{No curriculum dominates universally.} The relative advantage of forward versus reverse CL depends jointly on model reasoning capacity, task complexity, and the chosen difficulty metric. Stronger models benefit from forward CL on simpler tasks, while weaker models or harder tasks often favor reverse ordering—a pattern that holds across multiple metric families.

\noindent \circled{3}~\textbf{Curricula reshape optimization trajectories.} \textit{Inner-state} curricula primarily influence learning dynamics by adjusting convergence speed and stability, whereas \textit{task-aligned} curricula affect the final convergence point, shaping the model’s representations and generalization, suggesting that curricula can steer not only \textit{how} models learn, but also \textit{where} they ultimately converge.

\noindent \circled{4}~\textbf{Curricula modulate internal model states.} Beyond reasoning accuracy, CL shapes predictive uncertainty and confidence: Forward CL yields more cautious, uncertainty-aware models, while Reverse CL preserves decisiveness with higher confidence, highlighting that curricula influence not only \textit{what} models learn, but also \textit{how} they calibrate and express their internal properties.

Together, these results establish a unified framework for analyzing curriculum effects in LLMs and provide actionable guidance across different scenarios. For example, prioritizing decision-uncertain or high-confidence problems during training can further enhance a model’s reasoning capabilities.


\section{Method}
\label{sec:method}

\begin{figure}[htbp]
  \centering
  \includegraphics[width=\linewidth]{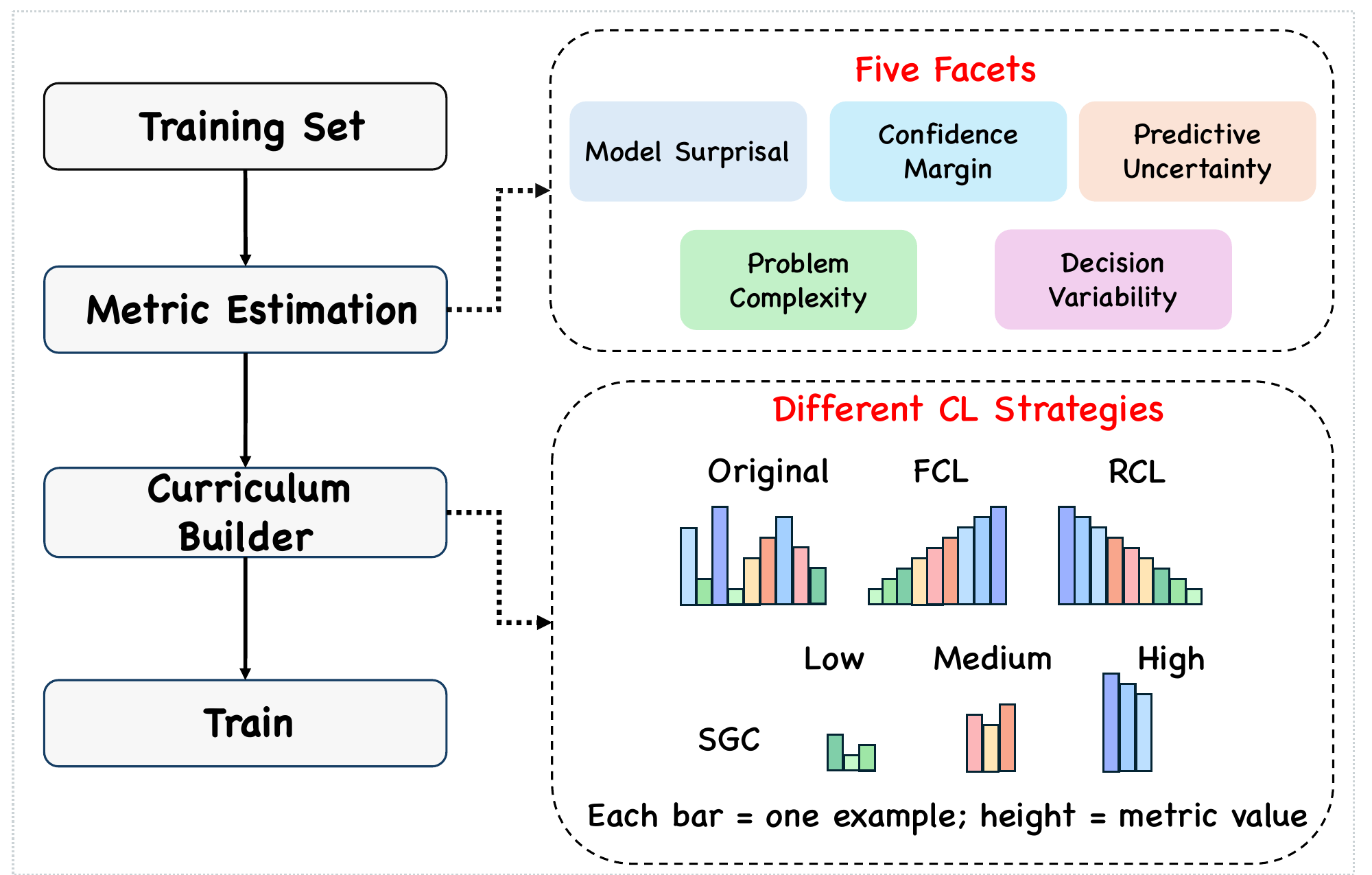}
    \caption{
    Overview of the metric-driven curriculum framework. 
    \textbf{Left:} From the training set, we estimate difficulty metrics, construct curricula, and train models under controlled settings. 
    \textbf{Right (top):} Five metric families spanning problem-side and model-side perspectives. 
    \textbf{Right (bottom):} Curriculum schedules, including \textit{Original} (random order), \textit{FCL} (easy$\rightarrow$hard), \textit{RCL} (hard$\rightarrow$easy), and \textit{SGC} (single-tier focus). 
    Each bar denotes a training instance, with height reflecting its metric value.
    }
  \label{fig:fig1}
\end{figure}

We design a metric-driven offline curriculum framework to systematically evaluate how different difficulty metrics and ordering strategies affect LLM reasoning, as shown in Fig~\ref{fig:fig1}. Our approach consists of three stages: \textit{(i)} quantifying difficulty through complementary problem-side and model-side metrics, \textit{(ii)} constructing curricula by reordering training data according to these metrics, and \textit{(iii)} evaluating the resulting models on both in-distribution and out-of-distribution benchmarks. Figure~\ref{fig:fig1} illustrates the complete pipeline.

\subsection{Difficulty Metrics: Two Complementary Views}
\label{sec:metric-definition}

To capture the multifaceted nature of difficulty, we decompose it into five metric families spanning two perspectives: \textit{problem-side metrics} that quantify intrinsic task complexity, and \textit{model-side metrics} that reflect the model's internal state when processing each example.

\subsubsection{Problem-Side Metrics}
\label{sec:p_metric}

Problem-side metrics assess difficulty independently of model behavior, either through human annotation or automated analysis. We collectively refer to these as \textit{Problem Difficulty} and consider four instantiations:

\noindent \textbf{Reasoning Step (RS)} counts the number of distinct logical operations required to reach a solution, capturing the depth of the reasoning chain.

\noindent \textbf{Symbol Complexity (SC)} quantifies notational richness on a 1–5 scale, where 1 denotes basic arithmetic and 5 denotes advanced constructs (integrals, limits, set theory).

\noindent \textbf{Comprehension Difficulty (CD)} measures how hard a problem is to \textit{understand} rather than solve, rated 1–5 based on ambiguity, context requirements, and prerequisite knowledge.

\noindent \textbf{Acc@\texorpdfstring{$K$}{K} (ACC)} quantifies empirical difficulty by sampling $K$ independent generations and computing the proportion of correct solutions:
\begin{equation}
\label{eq:acc_at_k}
\mathrm{ACC} = \frac{1}{K}\sum_{s=1}^{K} z_s = \hat{p},
\end{equation}
where $z_s \in {0,1}$ denotes the correctness of the $s$-th generation (1 if correct, 0 otherwise).

We estimate RS, SC, and CD using \textbf{Qwen2.5-Math-72B-Instruct}, while ACC is computed with respect to the target model under evaluation.

\subsubsection{Model-Side Metrics}
\label{model-metric}
Model-side metrics capture difficulty from the model's perspective by measuring internal properties during generation. To account for stochasticity, we decode with temperature 0.7 and average each metric over 20 independent completions per problem.

\paragraph{Notation} Given a problem $x$, model $M_\theta$ generates response $y_{1:T}$ token-by-token with distribution $p_\theta(y_t \mid y_{<t}, x)$. Let $\ell_t = \log p_\theta(y_t \mid y_{<t}, x)$ denote the log-probability of the emitted token. For the top-$k$ candidates at position $t$, we define the truncated distribution $q_{t,i} = e^{\ell_{t,i}}/\sum_{j=1}^{k} e^{\ell_{t,j}}$ (with $k=5$). We denote sorted log-probabilities as $\ell_{t,(1)} \ge \ell_{t,(2)} \ge \cdots$.

\paragraph{Model Surprisal} quantifies how unexpected a sequence is to the model via perplexity. We measure this at two granularities:

\textit{Sequence-Level Perplexity (SLP)} provides a holistic measure over the entire sequence:
\begin{equation}
\label{eq:sentence_ppl}
\text{SLP} = \exp\left(-\frac{1}{T} \sum_{t=1}^{T} \ell_t\right).
\end{equation}

\textit{Token-Level Perplexity (TLP)} captures local uncertainty by averaging token-wise entropy:
\begin{equation}
\label{eq:ppl_from_mean_entropy}
\text{TLP} = \exp\left( \frac{1}{T} \sum_{t=1}^{T} \left(-\sum_{i=1}^{k} q_{t,i}\log q_{t,i}\right) \right).
\end{equation}

\paragraph{Confidence Margin} assesses decisiveness by measuring the separation between top predictions. The \textit{Logit Gap (LG)} averages this margin across positions where at least two candidates exist:
\begin{equation}
\label{eq:logit_gap_avg_correct}
\mathrm{LG} = \frac{1}{U}\sum_{t \in \mathcal{G}} \big(\ell_{t,(1)} - \ell_{t,(2)}\big),
\end{equation}
where $\mathcal{G} = \{t : m_t \ge 2\}$ and $U = |\mathcal{G}|$.

\paragraph{Predictive Uncertainty} quantifies distributional dispersion via entropy. We compute this at both sequence and token levels:

\textit{Sequence-Level Entropy (SLE)} sums token-wise entropies:
\begin{equation}
\label{eq:sequence_entropy_trunc}
\mathrm{SLE} = \sum_{t=1}^{T} \left( - \sum_{i=1}^{k} q_{t,i}\,\log_{2} q_{t,i} \right).
\end{equation}

\textit{Token-Level Entropy (TLE)} measures local uncertainty:
\begin{equation}
\label{eq:token_entropy_trunc_single}
\mathrm{TLE} = - \sum_{i=1}^{k} q_{t,i}\,\log_{2} q_{t,i}.
\end{equation}

\paragraph{Decision Variability} measures the stability of a model’s predictions across repeated generations of the same problem. It is quantified by the \textit{Variance of Acc@\texorpdfstring{$K$}{K}} (VACC), which reflects fluctuations in correctness across $K$ trials. Given binary correctness indicators $z_s \in \{0,1\}$, we define
\begin{equation}
\label{eq:v_acc_at_k}
\mathrm{VACC} \;=\; \frac{1}{K}\sum_{s=1}^{K}\!\bigl(z_s - \hat{p}\bigr)^2 \;=\; \hat{p}\,\bigl(1-\hat{p}\bigr),
\end{equation}
where $\hat{p}$ is the empirical accuracy from Eq.~\ref{eq:acc_at_k}.

\subsection{Curriculum Construction}
\label{subsec:curr_construction}

Given a difficulty metric, we construct three primary curriculum strategies by reordering the training data:

\paragraph{Forward Curriculum Learning (FCL)} orders examples in ascending difficulty (easy-to-hard), allowing models to build foundational knowledge before tackling complex instances.

\paragraph{Reverse Curriculum Learning (RCL)} orders examples in descending difficulty (hard-to-easy), exposing models to challenging cases early to potentially accelerate learning.

\paragraph{Single Group Curriculum (SGC)} isolates a specific difficulty tier (Low, Medium, or High) by training exclusively on shuffled examples from that group, enabling controlled analysis of tier-specific effects.

We also explore two group based variants, \textbf{Group Forward Curriculum (GFC)} and \textbf{Group Reverse Curriculum (GRC)}, which partition the data into three tiers, shuffle samples within each tier, and then arrange the tiers in forward or reverse order. These strategies balance global difficulty progression with robustness to fine grained ranking noise. Full implementation details and additional experimental results are provided in Appendix~\ref{sec:additional-results}.

\subsection{Metric-driven Curriculum Framework}

For each combination of metric and curriculum strategy, we fine-tune a base model and evaluate it across multiple benchmarks to examine three questions: 
\textit{(i)} whether curriculum-based ordering improves over random shuffling, 
\textit{(ii)} which direction of progression (forward or reverse) is more effective, and 
\textit{(iii)} how these effects vary with model capacity and task complexity. 
In this offline setup, where the data remain fixed but their order is rearranged, we isolate the impact of curriculum structure while keeping the training objective, budget, and optimization procedure constant. 
This controlled design enables clear attribution of performance differences to data ordering alone.

\section{Experiment}
\subsection{Setup}
\paragraph{Datasets}
During training, we use MetaMathQA-40K~\cite{yu2023metamath} as our training dataset. It consists of 40K automatically generated, high-quality mathematical question--answer pairs specifically designed to evaluate and enhance advanced reasoning in large language models. The full MetaMathQA corpus is first shuffled, and 20K samples are randomly selected as the training set. To better evaluate generalization, we assess model performance on the in-distribution \textit{MetaMathQA (MMQA)} and four out-of-distribution mathematical reasoning benchmarks: \textit{ASDiv}~\cite{miao-etal-2020-diverse}, \textit{GSM8K}~\cite{cobbe2021training}, \textit{MATH}~\cite{hendrycks2021measuring}, and \textit{MathBench (MBench)}~\cite{liu2024mathbench}. For \textit{MBench}, we focus on application-style tasks from the \textit{Arithmetic}, \textit{Primary}, and \textit{Middle} difficulty levels.

\paragraph{Models}
We employ three open-source large language models in our experiments: Gemma3-4B-pt~\cite{team2025gemma}, Mistral-7B-v0.3~\cite{jiang2023mistral7b}, and Llama3.1-8B~\cite{dubey2024llama}. For brevity, we hereafter denote them as \textit{Gemma3-4B}, \textit{Llama3-8B}, and \textit{Mistral-7B}, respectively. All three are base (non-instruction-tuned) models, allowing us to isolate the effects of CL from instruction tuning or other external factors. Experimental configurations are described in Section~\ref{exp-details}.

\subsection{FCL vs. RCL with Multidimensional Metrics}
\label{exp:exp-1}

We evaluate Gemma3-4B, Llama3-8B, and Mistral-7B under different curriculum learning strategies on five mathematical reasoning benchmarks: four out-of-distribution datasets (ASDiv, GSM8K, MATH, and MBench) and one in-distribution dataset (MMQA). The in-distribution setting measures how curriculum ordering affects performance within the training domain, while the out-of-distribution setting evaluates generalization under distribution shift. Each dataset contains 512 randomly sampled problems. Results are reported across five metric dimensions: \textit{problem difficulty}, \textit{model surprisal}, \textit{confidence margin}, \textit{predictive uncertainty}, and \textit{decision variability}.

\subsubsection{Problem Difficulty}

Problem difficulty metrics include reasoning steps~(RS), symbol complexity (SC), comprehension difficulty (CD), and Acc@\texorpdfstring{$K$}{K} (ACC). 
Results on the five mathematical reasoning benchmarks are shown in Table~\ref{tab-1:problem-difficulty-cl}.

Overall, three consistent patterns emerge.~\textbf{(i)}~Curricula based on problem difficulty metrics consistently outperform the randomly shuffled baseline, with gains of up to 3--6\% on easier datasets such as \textit{ASDiv} and \textit{GSM8K} (e.g., \textit{Llama-8B}: +4.7\% on \textit{ASDiv} under RS$_{\downarrow}$), confirming that difficulty-aware ordering enhances reasoning performance.  \textbf{(ii)}~The benefit diminishes as task complexity increases: on \textit{MBench} and \textit{MATH}, improvements drop below 2\% or even reverse (e.g., \textit{Mistral-7B}: 18.8$\rightarrow$16.4\% on \textit{MATH}), suggesting that complex reasoning leaves less room for curriculum effects.   \textbf{(iii)}~Under distribution shift, Reverse CL generally excels on simpler datasets, whereas Forward CL performs better on harder ones (e.g., \textit{Gemma-4B}: 19.5$\rightarrow$24.8\% on \textit{MATH} under SC$_{\uparrow}$). Notably, \textit{Mistral-7B} consistently favors Reverse CL across all benchmarks, and reasoning-step and symbol-complexity metrics yield the most reliable gains.

\begin{table}[htbp]
\centering
\scriptsize
\setlength{\tabcolsep}{1.5pt}
\renewcommand{\arraystretch}{0.8}

\resizebox{\columnwidth}{!}{%
\begin{tabular}{@{} C{0.12\columnwidth}| C{0.12\columnwidth} | C{0.12\columnwidth}| *{4}{C{0.12\columnwidth}} @{}}

\toprule

\textbf{Model} & \textbf{Strategy} &
\textbf{MMQA} & \textbf{ASDiv} & \textbf{GSM8K} & \textbf{MBench} & \textbf{MATH} \\
\midrule
\multirow{9}{*}{\rotatebox[origin=c]{90}{\textbf{Gemma-4B}}}
 & Baseline         & 51.17 & 73.44 & 53.52 & 35.35 & 21.68 \\
 \cmidrule(l{-0pt}r{-0pt}){2-7}
 & RS$_{\uparrow}$   & 54.49 & 75.78 & 54.30 & 35.77 & 25.78 \\
 & RS$_{\downarrow}$ & 53.91 & 75.39 & 56.05 & 30.27 & 19.53 \\
 & SC$_{\uparrow}$   & 52.73 & 76.37 & 55.86 & 37.50 & 24.80 \\
 & SC$_{\downarrow}$ & 51.37 & 77.34 & 53.52 & 30.27 & 24.02 \\
 & CD$_{\uparrow}$   & 48.63 & 76.17 & 52.34 & 33.01 & 21.88 \\
 & CD$_{\downarrow}$ & 50.59 & 76.56 & 54.10 & 32.42 & 20.90 \\
 & ACC$_{\uparrow}$  & 58.59 & 75.20 & 56.25 & 28.32 & 24.61 \\
 & ACC$_{\downarrow}$& 56.64 & 77.15 & 57.42 & 36.52 & 23.63 \\
\midrule
\multirow{9}{*}{\rotatebox[origin=c]{90}{\textbf{Mistral-7B}}}
 & Baseline          & 54.30 & 68.55 & 60.35 & 36.72 & 18.75 \\
 \cmidrule(l{-0pt}r{-0pt}){2-7}
 & RS$_{\uparrow}$   & 50.59 & 74.22 & 59.77 & 36.91 & 16.41 \\
 & RS$_{\downarrow}$ & 60.16 & 73.24 & 65.62 & 38.09 & 17.58 \\
 & SC$_{\uparrow}$   & 51.37 & 71.68 & 55.66 & 39.45 & 16.41 \\
 & SC$_{\downarrow}$ & 50.59 & 72.85 & 59.57 & 37.89 & 16.99 \\
 & CD$_{\uparrow}$   & 52.15 & 72.85 & 57.81 & 36.72 & 16.21 \\
 & CD$_{\downarrow}$ & 53.76 & 73.63 & 62.30 & 38.12 & 16.80 \\
 & ACC$_{\uparrow}$  & 47.27 & 63.87 & 54.30 & 35.55 & 14.65 \\
 & ACC$_{\downarrow}$& 52.15 & 71.48 & 60.35 & 36.52 & 16.80 \\
\midrule
\multirow{9}{*}{\rotatebox[origin=c]{90}{\textbf{Llama-8B}}}
 & Baseline         & 59.57 & 74.02 & 62.50 & 38.09 & 18.55 \\
\cmidrule(l{-0pt}r{-0pt}){2-7}
 & RS$_{\uparrow}$   & 54.49 & 77.34 & 60.35 & 35.74 & 21.88 \\
 & RS$_{\downarrow}$ & 63.67 & 77.73 & 70.90 & 38.67 & 20.31 \\
 & SC$_{\uparrow}$   & 61.72 & 75.78 & 64.95 & 38.48 & 20.90 \\
 & SC$_{\downarrow}$ & 59.79 & 74.61 & 64.85 & 37.70 & 17.77 \\
 & CD$_{\uparrow}$   & 49.61 & 71.09 & 57.33 & 31.64 & 19.74 \\
 & CD$_{\downarrow}$ & 52.93 & 76.95 & 62.89 & 36.33 & 19.13 \\
 & ACC$_{\uparrow}$  & 56.25 & 76.17 & 60.16 & 36.91 & 21.02 \\
 & ACC$_{\downarrow}$& 55.86 & 78.71 & 65.04 & 36.52 & 20.51 \\
\bottomrule
\end{tabular}%
}
\caption{Accuracy (\%) on five mathematical reasoning benchmarks for curricula constructed from \textbf{problem difficulty} metrics (RS, SC, CD, ACC). The \textbf{baseline} uses randomly shuffled data; $\uparrow$ and $\downarrow$ indicate ascending (Forward CL) and descending (Reverse CL) ordering, respectively. See Sec.~\ref{sec:metric-definition} for metric definitions.}

\label{tab-1:problem-difficulty-cl}
\end{table}


\subsubsection{Model Surprisal}

Model surprisal is quantified using sequence-level perplexity (SLP) and token-level perplexity (TLP). Results are reported in Table~\ref{tab-2:perplexity-cl}.

We can observe that perplexity serves as an effective yet partially unstable curriculum signal for mathematical reasoning. \textbf{(i)}~In most cases, Forward CL (low to high perplexity) outperforms Reverse CL, suggesting that starting from low-perplexity samples provides a more stable optimization signal, consistent with prior findings~\cite{zhang2025beyond}. \textbf{(ii)}~However, the directionality is not universal. In some cases, Reverse CL achieves higher gains (e.g., \textit{Mistral-7B}: +2.7\% on \textit{GSM8K} under SLP$_{\downarrow}$), implying that high-perplexity samples can help models adapt to semantic noise and improve robustness~\cite{zhang2024noise}, while Forward CL may overfit early confident patterns. 
\textbf{(iii)}~SLP provides a more stable signal for curriculum direction, favoring Forward CL in most cases (11/15, including all for \textit{Gemma-4B} and \textit{Llama-8B}). In contrast, TLP exhibits less stability but often achieves stronger gains, with RCL frequently surpassing FCL on harder datasets (e.g., \textit{Llama-8B}: 21.09$\rightarrow$24.02 on \textit{MATH}), suggesting its heightened sensitivity to local variations can occasionally yield larger improvements.

\begin{table}[htbp]
\centering
\scriptsize
\setlength{\tabcolsep}{1.5pt}
\renewcommand{\arraystretch}{0.9}

\resizebox{\columnwidth}{!}{%
\begin{tabular}{@{} C{0.12\columnwidth}| C{0.12\columnwidth}| C{0.12\columnwidth}| *{4}{C{0.12\columnwidth}} @{}}
\toprule
\textbf{Model} & \textbf{Strategy} &
\textbf{MMQA} & \textbf{ASDiv} & \textbf{GSM8K} & \textbf{MBench} & \textbf{MATH} \\

\midrule

\multirow{5}{*}{\rotatebox[origin=c]{90}{\textbf{Gemma-4B}}}
 & Baseline       & 51.17 & 73.44 & 53.52 & 35.35 & 21.68 \\
 \cmidrule(l{-0pt}r{-0pt}){2-7}
 & SLP$_{\uparrow}$ & 56.84 & 75.20 & 56.84 & 32.42 & 24.41 \\
 & SLP$_{\downarrow}$ & 48.63 & 73.44 & 52.15 & 31.45 & 24.02 \\
 & TLP$_{\uparrow}$ & 53.12 & 75.59 & 54.30 & 38.48 & 22.85 \\
 & TLP$_{\downarrow}$ & 46.09 & 71.88 & 49.02 & 33.79 & 20.70 \\
\midrule

\multirow{5}{*}{\rotatebox[origin=c]{90}{\textbf{Mistral-7B}}}
 & Baseline       & 54.30 & 68.55 & 60.35 & 36.72 & 18.75 \\
 \cmidrule(l{-0pt}r{-0pt}){2-7}
 & SLP$_{\uparrow}$ & 51.17 & 72.27 & 57.81 & 40.23 & 17.38 \\
 & SLP$_{\downarrow}$ & 52.73 & 73.24 & 63.09 & 40.43 & 17.19 \\
 & TLP$_{\uparrow}$ & 53.32 & 70.70 & 62.70 & 40.04 & 18.55 \\
 & TLP$_{\downarrow}$ & 54.49 & 73.24 & 60.55 & 39.43 & 16.21 \\
\midrule

\multirow{5}{*}{\rotatebox[origin=c]{90}{\textbf{Llama-8B}}}
 & Baseline       & 59.57 & 74.02 & 62.50 & 38.09 & 18.55 \\
 \cmidrule(l{-0pt}r{-0pt}){2-7}
 & SLP$_{\uparrow}$ & 57.23 & 76.76 & 65.43 & 42.04 & 20.90 \\
 & SLP$_{\downarrow}$ & 50.98 & 72.66 & 60.16 & 40.04 & 18.36 \\
 & TLP$_{\uparrow}$ & 58.40 & 77.15 & 67.77 & 33.20 & 21.09 \\
 & TLP$_{\downarrow}$ & 58.98 & 75.39 & 66.02 & 36.72 & 24.02 \\
\bottomrule
\end{tabular}%
}
\caption{Accuracy (\%) on five mathematical reasoning benchmarks under curricula constructed with \textbf{model surprisal} (SLP, TLP).}
\label{tab-2:perplexity-cl}
\end{table}

\subsubsection{Confidence Margin}

Confidence margin, measured by the logit gap~(LG), quantifies how confidently a model prefers its top prediction over alternatives. Larger gaps indicate stronger confidence in token-level~decisions.

Table~\ref{tab:lg_cl_evaluation} shows that the logit gap (LG) serves as an effective and stable curriculum signal for mathematical reasoning. 
\textbf{(i)}~Across models, LG-based curricula consistently outperform the shuffled baseline, particularly on easier datasets such as \textit{ASDiv} (e.g., \textit{Llama-8B}: 74.02$\rightarrow$77.93\%), following a consistent trend (\textsc{RCL} $>$ Baseline $>$ \textsc{FCL}). This confirms that LG provides a reliable measure of task difficulty and model confidence. 
\textbf{(ii)}~Reverse CL (high-confidence first) strengthens model confidence by exposing it to high-margin examples early in training, allowing the model to form more decisive predictions and improving reasoning performance (e.g., \textit{Gemma-4B}: +3.5\% on \textit{ASDiv}, +3.0\% on \textit{MATH}). Overall, LG offers a clear and direction-sensitive curriculum signal, with Reverse CL providing the most consistent gains across benchmarks.

\begin{table}[htbp]
\centering
\scriptsize
\setlength{\tabcolsep}{1.5pt}
\renewcommand{\arraystretch}{1.2}

\begin{tabular}{@{} C{0.12\columnwidth}| C{0.12\columnwidth}| C{0.12\columnwidth}| *{4}{C{0.12\columnwidth}} @{}}
\toprule
\textbf{Model} & \textbf{Strategy} &
\textbf{MMQA} & \textbf{ASDiv} & \textbf{GSM8K} & \textbf{MBench} & \textbf{MATH} \\
\midrule

\multirow{3}{*}{\rotatebox[origin=c]{90}{\textbf{Gemma-4B}}}
 & Baseline            & 51.17 & 73.44 & 53.52 & 35.35 & 21.68 \\
 \cmidrule(l{-0pt}r{-0pt}){2-7}
 & LG$_{\uparrow}$   & 52.73 & 75.20 & 53.52 & 35.55 & 24.61 \\
 & LG$_{\downarrow}$ & 54.30 & 77.15 & 57.03 & 38.09 & 24.72 \\
\midrule

\multirow{3}{*}{\rotatebox[origin=c]{90}{\textbf{Mistral-7B}}}
 & Baseline             & 54.30 & 68.55 & 60.35 & 36.72 & 18.75 \\
 \cmidrule(l{-0pt}r{-0pt}){2-7}
 & LG$_{\uparrow}$   & 55.27 & 74.22 & 60.16 & 40.04 & 16.21 \\
 & LG$_{\downarrow}$ & 52.24 & 70.12 & 59.96 & 40.23 & 17.97 \\
\midrule

\multirow{3}{*}{\rotatebox[origin=c]{90}{\textbf{Llama-8B}}}
 & Baseline            & 59.57 & 74.02 & 62.50 & 38.09 & 18.55 \\
 \cmidrule(l{-0pt}r{-0pt}){2-7}
 & LG$_{\uparrow}$   & 56.25 & 77.93 & 69.92 & 34.77 & 19.73 \\
 & LG$_{\downarrow}$ & 58.79 & 75.39 & 66.80 & 38.48 & 21.68 \\
\bottomrule

\end{tabular}

\caption{Accuracy (\%) on five mathematical reasoning benchmarks under curricula constructed with \textbf{confidence margin} (LG).}
\label{tab:lg_cl_evaluation}
\end{table}

\subsubsection{Predictive Uncertainty}
The predictive uncertainty of a model is measured by entropy-based metrics, including entropy at the sentence level (SLE) and entropy at the token level (TLE). These metrics reflect how dispersed or peaked the model’s predictive distributions are, with higher entropy indicating greater uncertainty in its predictions. 

\begin{table}[htbp]
\centering
\scriptsize
\setlength{\tabcolsep}{2pt}
\renewcommand{\arraystretch}{1.05}

\begin{tabular}{@{} C{0.12\columnwidth}| C{0.12\columnwidth}| C{0.12\columnwidth}| *{4}{C{0.12\columnwidth}} @{}}
\toprule
\textbf{Model} & \textbf{Strategy} &
\textbf{MMQA} & \textbf{ASDiv} & \textbf{GSM8K} & \textbf{MBench} & \textbf{MATH} \\
\midrule

\multirow{5}{*}{\rotatebox[origin=c]{90}{\textbf{Gemma-4B}}}
 & Baseline        & 51.17 & 73.44 & 53.52 & 35.35 & 21.68 \\
 \cmidrule(l{-0pt}r{-0pt}){2-7}
 & SLE$_{\uparrow}$ & 53.71 & 75.59 & 55.86 & 34.57 & 24.22 \\
 & SLE$_{\downarrow}$ & 53.52 & 78.52 & 55.47 & 33.40 & 23.63 \\
 & TLE$_{\uparrow}$ & 53.52 & 75.78 & 56.05 & 35.16 & 22.85 \\
 & TLE$_{\downarrow}$ & 50.59 & 75.59 & 53.91 & 31.45 & 25.79 \\
\midrule

\multirow{5}{*}{\rotatebox[origin=c]{90}{\textbf{Mistral-7B}}}
 & Baseline        & 54.30 & 68.55 & 60.35 & 36.72 & 18.75 \\
 \cmidrule(l{-0pt}r{-0pt}){2-7}
 & SLE$_{\uparrow}$ & 54.49 & 72.27 & 59.57 & 38.87 & 14.84 \\
 & SLE$_{\downarrow}$ & 52.73 & 74.02 & 59.96 & 39.94 & 18.75 \\
 & TLE$_{\uparrow}$ & 54.66 & 70.70 & 60.16 & 40.21 & 16.80 \\
 & TLE$_{\downarrow}$ & 50.00 & 73.24 & 60.74 & 40.43 & 16.98 \\
\midrule

\multirow{5}{*}{\rotatebox[origin=c]{90}{\textbf{Llama-8B}}}
 & Baseline        & 59.57 & 74.02 & 62.50 & 38.09 & 18.55 \\
 \cmidrule(l{-0pt}r{-0pt}){2-7}
 & SLE$_{\uparrow}$ & 58.98 & 79.49 & 61.91 & 39.45 & 21.29 \\
 & SLE$_{\downarrow}$ & 56.25 & 74.02 & 62.89 & 34.96 & 21.88 \\
 & TLE$_{\uparrow}$ & 58.01 & 72.85 & 70.12 & 35.94 & 23.83 \\
 & TLE$_{\downarrow}$ & 55.47 & 76.95 & 56.84 & 36.72 & 22.46 \\
\bottomrule

\end{tabular}


\caption{Accuracy (\%) on five mathematical reasoning benchmarks with curricula derived from \textbf{predictive uncertainty} (entropy-based metrics, SLE and TLE).}
\label{tab:es_et_cl}
\end{table}

Table~\ref{tab:es_et_cl} shows that entropy-based curricula provide informative yet unstable difficulty signals. 
\textbf{(i)}~Both SLE and TLE effectively capture predictive uncertainty, especially on easier datasets such as \textit{ASDiv} (\textit{Llama-8B}: 79.49\% under SLE$_{\uparrow}$). However, their stability is lower than that of logit-gap or perplexity-based metrics, and performance occasionally declines on harder tasks (\textit{Mistral-7B}: –3.9\% on \textit{MATH}). 
\textbf{(ii)}~For in-distribution evaluation, forward CL consistently outperforms reverse CL across models, indicating that ordering by confidence (low-entropy first) facilitates more stable optimization. 
\textbf{(iii)}~Between metrics, TLE is more noise-prone yet occasionally yields larger gains (\textit{Llama-8B}: +7.6\% on \textit{MBench} under TLE$_{\uparrow}$), consistent with findings that token-level entropy is often dominated by noisy high-entropy outliers rather than stable uncertainty patterns~\cite{wang2025beyond}. In contrast, SLE aggregates uncertainty more coherently, offering steadier, model-agnostic improvements. Overall, entropy-based curricula provide informative but unstable signals.

\subsubsection{Decision Variability}
Decision variability, measured as the Variance of Acc@\texorpdfstring{$K$}{K} (VACC). It captures how “blurred” a problem appears to the model by quantifying the variability of correctness across $K$ independent generations for the same problem.

\begin{table}[htbp]
\centering
\scriptsize
\setlength{\tabcolsep}{2pt}
\renewcommand{\arraystretch}{1.2}

\begin{tabular}{@{} C{0.12\columnwidth}| C{0.12\columnwidth}| C{0.12\columnwidth}| *{4}{C{0.12\columnwidth}} @{}}
\toprule
\textbf{Model} & \textbf{Strategy} &
\textbf{MMQA} & \textbf{ASDiv} & \textbf{GSM8K} & \textbf{MBench} & \textbf{MATH} \\
\midrule

\multirow{3}{*}{\rotatebox[origin=c]{90}{\textbf{Gemma-4B}}}
 & Baseline            & 51.17 & 73.44 & 53.52 & 35.35 & 21.68 \\
 \cmidrule(l{-0pt}r{-0pt}){2-7}
 & VACC$_{\uparrow}$ & 56.25 & 76.56 & 57.42 & 31.45 & 26.17 \\
 & VACC$_{\downarrow}$ & 60.16 & 75.39 & 58.40 & 36.72 & 24.41 \\
\midrule

\multirow{3}{*}{\rotatebox[origin=c]{90}{\textbf{Mistral-7B}}}
 & Baseline            & 54.30 & 68.55 & 60.35 & 36.72 & 18.75 \\
 \cmidrule(l{-0pt}r{-0pt}){2-7}
 & VACC$_{\uparrow}$ & 47.27 & 63.87 & 54.30 & 35.55 & 14.65 \\
 & VACC$_{\downarrow}$ & 55.15 & 71.48 & 60.67 & 36.92 & 16.80 \\
\midrule

\multirow{3}{*}{\rotatebox[origin=c]{90}{\textbf{Llama-8B}}}
 & Baseline            & 59.57 & 74.02 & 62.50 & 38.09 & 18.55 \\
 \cmidrule(l{-0pt}r{-0pt}){2-7}
 & VACC$_{\uparrow}$ & 56.05 & 76.76 & 63.28 & 35.16 & 19.53 \\
 & VACC$_{\downarrow}$ & 63.09 & 78.91 & 60.97 & 41.02 & 21.48 \\
\bottomrule

\end{tabular}


\caption{Accuracy (\%) on five mathematical reasoning benchmarks with curricula derived from \textbf{decision variability} (VACC).}
\label{tab:vacc_cl}
\end{table}

Table~\ref{tab:vacc_cl} shows that variability-based curricula exhibit clear directional asymmetries. \textbf{(i)}~Reverse CL, which begins with high-variability (less stable) problems, consistently outperforms Forward CL across most settings—winning 12 of 15 comparisons—with notable gains on easier datasets (\textit{MMQA}: +5.9\%, \textit{ASDiv}: +4.9\% for \textit{Llama-8B}). 
\textbf{(ii)}~Forward CL is competitive only in isolated cases, such as \textit{Gemma-4B} on \textit{MATH} (+1.8\%), where uncertainty is already limited. \textbf{(iii)}~Across models, the performance gap between directions narrows on more difficult benchmarks (\textit{MBench}, \textit{MATH}), suggesting that once reasoning difficulty dominates, the influence of variability ordering diminishes. Overall, these results indicate that initiating learning with high-variability examples (Reverse CL) promotes faster adaptation on simpler reasoning tasks, whereas both strategies converge as task complexity increases.

\subsection{Metric-Tiered Curricula}
\label{exp:exp-2}

Building on the analysis of how different CL strategies influence model reasoning in the previous section, we further investigate \textit{how} samples with varying metric levels affect learning across tasks of different difficulty. To this end, we conduct \textbf{metric tiered training} on \textit{Llama3 8B}. For each metric, the training set is divided into three equal sized tiers, namely \textbf{Low}, \textbf{Medium}, and \textbf{High}, based on metric values. Within each tier, samples are shuffled, and the model is trained on one tier at a time under a fixed data budget to isolate the effect of each difficulty band. Evaluation is performed on four benchmarks of increasing complexity, including \textit{ASDiv}, \textit{GSM8K}, \textit{MBench}, and \textit{MATH}, which are further grouped into \textit{easy} and \textit{hard} categories to analyze tier effects across task difficulty. The five metrics (Section~\ref{sec:metric-definition}) represent complementary dimensions: problem difficulty (ACC), model surprisal (TLP), predictive uncertainty (TLE), confidence margin (LG), and decision variability (VACC). A \textbf{Baseline} trained on a randomly shuffled one third subset of the data serves as a size matched control.

\begin{figure}[htbp]
  \centering
  \includegraphics[width=\linewidth]{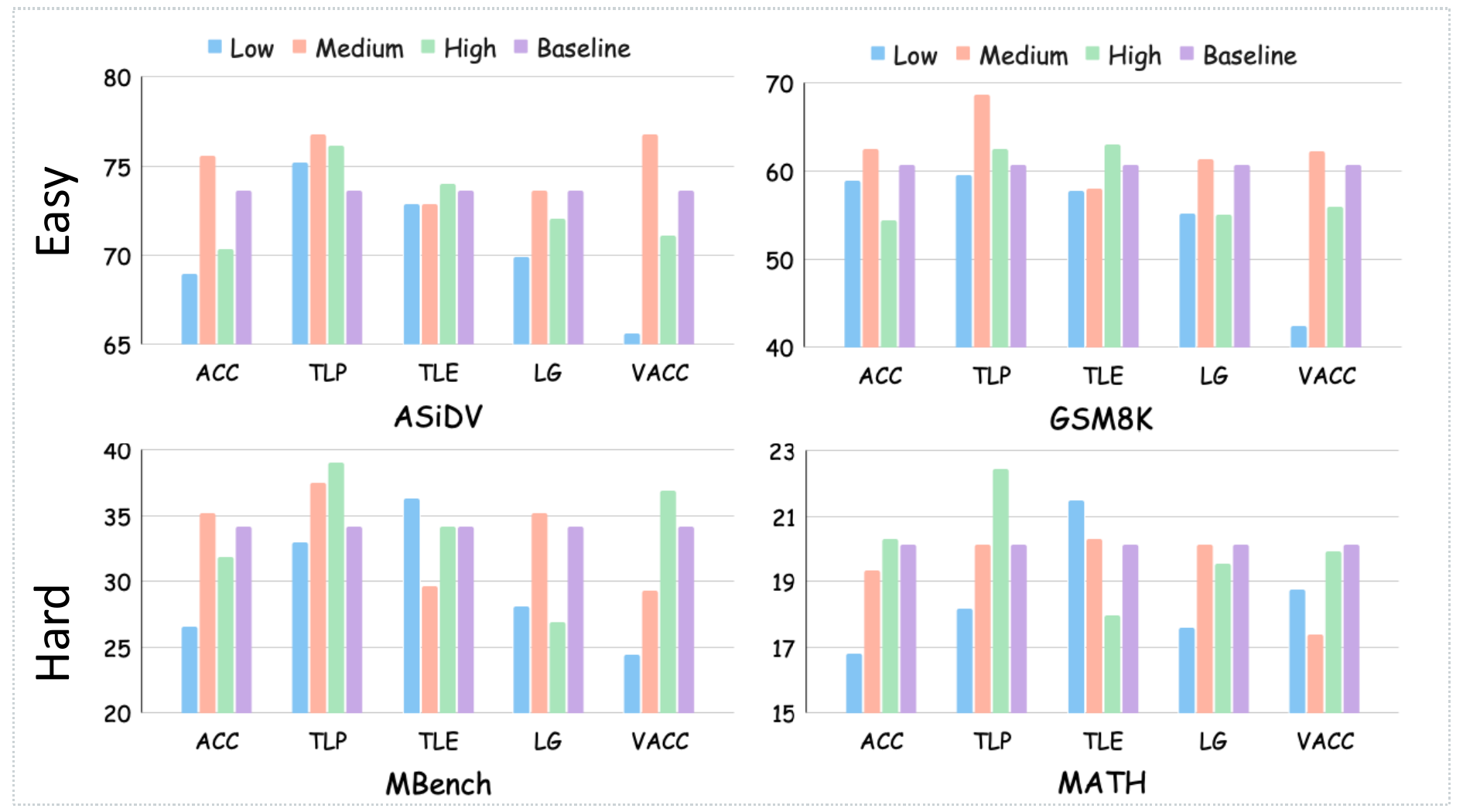}
  \caption{Results of metric-tiered curricula on Llama3-8B for four benchmarks. For each metric, the training set is split into three equal-size tiers (Low, Medium, High), each comprising one third of the data and shuffled; models are trained on one tier at a time. Bars show~accuracy (\%).}
  \label{fig:fig2}
\end{figure}

The results in Figure~\ref{fig:fig2} reveal two consistent trends. \textbf{(i)}~For entropy-based metrics, the most effective tier depends on task difficulty. On easier benchmarks (\textit{ASDiv}, \textit{GSM8K}), training on the \textbf{high-entropy} tier yields stronger gains, suggesting that uncertain samples provide valuable learning signals for simpler tasks. In contrast, on harder benchmarks (\textit{MBench}, \textit{MATH}), the \textbf{low-entropy} tier performs better, indicating that lower-uncertainty examples become more useful as task complexity increases. \textbf{(ii)}~For the remaining metrics (ACC, TLP, LG, VACC), improvements are largely concentrated in the \textbf{medium} tier. Task difficulty further modulates this effect: easier tasks benefit more from the low-to-medium range, while harder ones favor the medium-to-high range. This pattern suggests that low-surprisal, low-confidence, and low-variability examples support simpler reasoning, whereas challenging reasoning tasks rely more on informative, high-difficulty samples.

\subsection{Effect of CL Strategies on Internal Metrics}
\label{exp: internal metrics}

Beyond external performance, we further examine \textit{how curricula influence internal model states}, including confidence and uncertainty, and \textit{how the direction of the curriculum affects these properties}. We fine tune \textit{Llama3 8B} using three training schedules: Forward CL (FCL), Reverse CL (RCL), and a randomly shuffled baseline (SHUF). Each schedule is constructed using one of five metrics that represent different dimensions of difficulty: TLP, SLP, TLE, SLE, and LG (see Section~\ref{sec:metric-definition}). For each metric, the training data are reordered based on its values, and the resulting models are evaluated on five benchmarks \textit{using the same metric that guided training}. From each dataset, we sample 200 problems and report the average metric values.

\begin{figure*}[htbp]
  \centering
  \includegraphics[width=\linewidth]{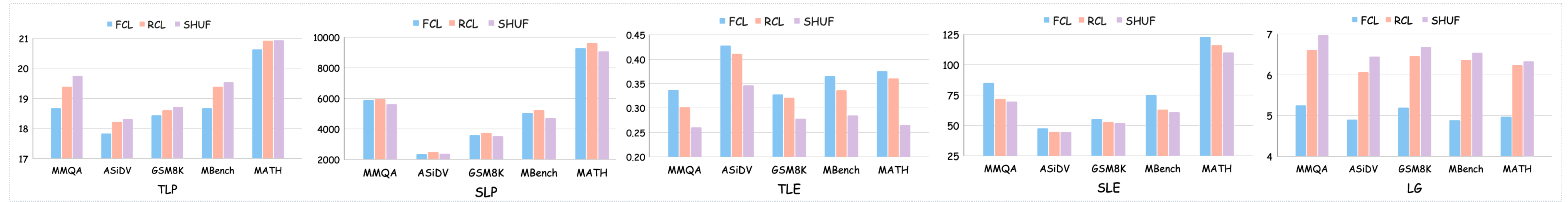}
    \caption{Internal-state metrics across different curriculum strategies. We present TLP, SLP, TLE, SLE, and LG (from left to right, defined in Section~\ref{sec:metric-definition}) for \textit{Llama3-8B} trained under Forward CL (FCL), Reverse CL (RCL), and the shuffled baseline (SHUF) across five benchmarks.}
  \label{fig:inner_state}
\end{figure*}

\begin{figure*}[htbp]
  \centering
  \includegraphics[width=\linewidth]{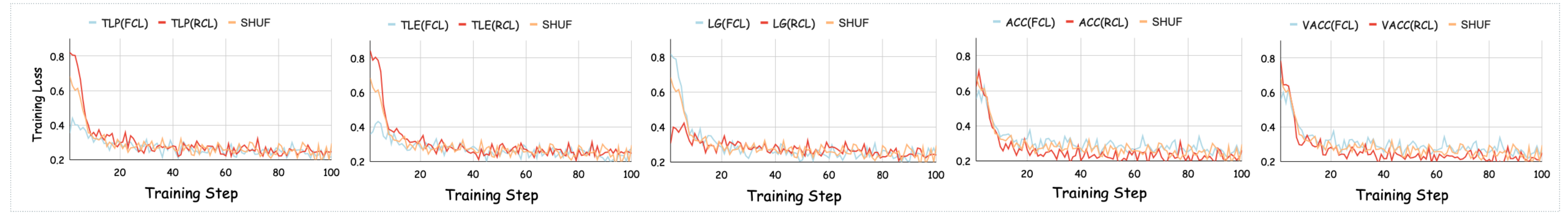}
\caption{
Training loss dynamics over 100 training steps under different curriculum learning strategies. 
We compare FCL, RCL, and the shuffled baseline (SHUF) across five metric dimensions (from left to right): TLP, TLE, LG, ACC, and VACC, showing how loss evolves over training under different CL strategies.
}

  \label{fig:train-loss}
\end{figure*}

The results are shown in Figure~\ref{fig:inner_state}. Across all five internal metrics, curriculum direction induces clear and consistent shifts in model internal states. Both FCL and RCL systematically move the model away from the randomized baseline (SHUF), confirming that curriculum ordering fundamentally reshapes the model’s internal confidence and uncertainty distributions. Specifically, for TLP, SLP, and LG, we observe a consistent trend of FCL $<$~RCL, suggesting that training on low-perplexity or low-confidence samples first yields more conservative and stable internal representations. In contrast, SLE shows the opposite behavior, where prioritizing low-certainty data increases model uncertainty, indicating a trade-off between calibration and robustness under different curriculum directions.

\subsection{Impact of CL Strategies on Training Dynamics}
In previous experiments, we demonstrated that CL strategies substantially influence both the model’s external performance and internal states. Here, we further investigate \textit{how different CL strategies shape the model’s training dynamics}. We consider five representative metrics spanning distinct learning dimensions: \textit{problem complexity (ACC)}, \textit{model surprisal (TLP)}, \textit{model uncertainty (TLE)}, \textit{confidence margin (LG)}, and \textit{decision variability (VACC)} (see Section~\ref{sec:metric-definition}). Training loss is tracked under two curriculum schedules: Forward CL (FCL) and Reverse CL (RCL), with a randomly shuffled curriculum (SHUF) serving as the baseline. This setup enables a systematic examination of how curriculum organization shapes model optimization behavior over time.

We present results in Figure~\ref{fig:train-loss}. These results reveal three distinct mechanisms through which curriculum strategies influence training dynamics. \textbf{\textit{(i) Internal-state metrics}} (TLP, TLE, LG) capture difficulty from the model’s own predictive behavior and primarily regulate the convergence dynamics during training: starting with less-perplexed, more-certain, or less-confident samples produces higher initial loss but faster convergence, while beginning with high-perplexity or high-confidence samples yields slower convergence yet similar final loss. Such metrics mainly shape the model’s internal uncertainty and confidence distributions, indirectly enhancing reasoning ability. \textbf{\textit{(ii) Task-aligned metrics}} (ACC, VACC) measure difficulty using task-level outcomes that reflect alignment with the learning objective, having limited effect on convergence speed but substantial impact on the final loss and generalization: training on easier or higher-variability samples leads to lower steady-state loss, indicating improved data fitting. Overall, curriculum learning influences training through two complementary pathways: \textit{internal-state metrics shape model properties such as confidence and uncertainty, while task-aligned metrics affect data fitting and generalization. Together, these mechanisms explain why CL strategies can effectively enhance reasoning performance.}

\section*{Conclusion}

We propose a multidimensional offline curriculum learning framework to systematically analyze how forward and reverse curricula affect LLMs in mathematical reasoning across three open-source base models: \textit{Gemma 4B}, \textit{Mistral 7B}, and \textit{Llama 8B}. Our results show that well-chosen CL strategies can significantly enhance reasoning performance, with the optimal direction depending jointly on model capability and task complexity. Across different difficulty metrics such as logit gap, entropy, and perplexity, the effectiveness of training tiers varies with task difficulty: harder tasks benefit from high-difficulty and high-uncertainty samples, whereas easier tasks favor low-difficulty and high-confidence ones. Moreover, CL influences not only external performance but also internal calibration and optimization dynamics. \textit{Internal-state metrics} influence model confidence and uncertainty, while \textit{task-aligned metrics} affect data fitting and generalization. Overall, these results highlight the key role of data organization in shaping learning dynamics and provide practical insights for designing effective curricula for LLM post-training.

\clearpage
\newpage
\section*{Limitations}
While this work systematically investigates curriculum learning (CL) strategies for LLMs, several limitations remain. \textbf{(i)} We focus on \textit{offline} CL, where data ordering is predefined before training. Although this enables controlled comparisons, it does not capture the adaptivity of \textit{online} or self-paced curricula that evolve dynamically during training; we leave this for future work. \textbf{(ii)} Our analysis mainly examines solution accuracy and internal model states. While these metrics reveal how CL influences learning dynamics, we have not yet explored the model’s full reasoning trajectories or intermediate solution steps, which we plan to investigate in future research.

\section*{Ethical Considerations}
All experiments presented in this study were conducted using publicly available datasets and models licensed for academic research purposes. To the best of our knowledge, this work does not present any ethical concerns.

\bibliography{ref.bib}

\newpage
\appendix

\clearpage          
\appendix

\startcontents[appendices]
\section*{Contents of Appendix}
\printcontents[appendices]{}{1}{\normalsize}

\newpage

\section{Related Work}

\subsection{Curriculum Learning for Large Language Models}

Curriculum learning (CL),  proposed in machine learning and cognitive science—trains models on examples arranged from easy to hard to stabilize optimization and improve efficiency~\cite{bengio2009curriculum, elman1993learning, kumar2010self}. This principle has recently been scaled to large language models (LLMs). Several works incorporate CL directly into LLM training: for instance, \citeauthor{zelikman2024star} construct an implicit easy-to-hard curriculum via self-training on verified chains of thought, boosting mathematical and symbolic reasoning; \citeauthor{huang2024it2acl} design a two-phase automated curriculum across tasks and within-task instructions, yielding stronger performance and better zero-shot generalization; and other researchers develop CL schemes driven by diverse difficulty metrics~\cite{chen2023skill, rao-etal-2024-commonit, xi2024training, parashar2025curriculum}. Complementing training-time CL, \citeauthor{zhou2023leasttomost} apply CL at test time by decomposing complex problems into ranked subproblems and solving them from easy to hard, markedly enhancing reasoning on challenging benchmarks. A parallel line of work applies CL before or alongside pretraining by reordering or sampling data to improve efficiency and downstream reasoning~\cite{zhang-etal-2025-preference, kim2024strategic, feng2023citing}. For example, \citeauthor{zhang-etal-2025-preference} cast preference-aware data selection as a pretraining-time curriculum, prioritizing high-utility examples and phasing in lower-utility data, which improves sample efficiency and downstream performance relative to uniform or domain-agnostic mixtures.

\subsection{Curriculum Metrics for Mathematical Reasoning}
Mathematical reasoning tasks naturally exhibit graded difficulty, making them well-suited for curriculum learning approaches that progress from easier to harder problems. Existing work on curriculum metrics for this domain can be broadly divided into two categories: problem-side metrics and model-conditional metrics. Problem-side metrics are derived from an instance’s structure and meaning, not from model behavior, and may be computed analytically or via annotation. For example, \citeauthor{jung2025reasoning} order instances by reasoning depth, using shallow-to-deep pacing to guide training, while \citeauthor{hammoud2025train} employ a length-based curriculum that begins with generous chain-of-thought budgets and gradually anneals them to encourage concise reasoning. Other studies define difficulty through symbolic complexity~\cite{tong2024dart}, linguistic complexity~\cite{srivatsa-kochmar-2024-makes}, or verifiability signals such as self-consistency and stepwise correctness~\cite{wang2022self,lightman2023let}, leveraging these properties to schedule learning. Model-conditional metrics instead adapt curricula based on model behavior. Perplexity has been used as a difficulty signal to sequence training data and in-context examples, yielding faster convergence and improved in-context learning performance~\cite{zhou2020uncertainty,liu2024let}. Entropy-based uncertainty has also been proposed as a pacing signal, where prioritizing high-information samples leads to more stable training and stronger downstream results, including higher BLEU scores~\cite{zhou2020uncertainty,fang2025entropy}. Beyond these, logit gap signals~\cite{jafarpour2021active}, calibrated confidence estimates~\cite{ao2023confidence}, and adaptive verification metrics~\cite{wang2022self,song2022adaptive} have been explored to construct easy-to-hard schedules that accelerate training efficiency and boost generalization.

Despite growing interest in curriculum learning, there is limited work that systematically compares curriculum metrics and examines how CL strategies relate to task type and model reasoning capabilities. To address this gap, we introduce a multidimensional evaluation benchmark for head-to-head comparisons of CL strategies. To control for confounds, we focus on offline curricula and conduct experiments on base language models (i.e., non–instruction-tuned language models).

\section{Additional Curriculum Strategies}
In addition to the Forward CL and Reverse CL settings introduced in Section~\ref{subsec:curr_construction}, we explore two group-based curriculum variants that aim to reduce the sensitivity of training to fine-grained ranking noise.

\paragraph{Group Forward Curriculum (GFC).}
The dataset is partitioned into three tiers—\textit{Low}, \textit{Medium}, and \textit{High}—based on metric values using thresholding or quantile splits. Samples within each tier are randomly shuffled to remove intra-group ordering effects. Training proceeds from easier to harder groups (\textit{Low} $\rightarrow$ \textit{Medium} $\rightarrow$ \textit{High}), preserving the forward progression of difficulty.

\paragraph{Group Reverse Curriculum (GRC).}
Using the same grouping procedure, training is instead conducted in reverse order (\textit{High} $\rightarrow$ \textit{Medium} $\rightarrow$ \textit{Low}). This variant emphasizes difficult samples early while maintaining stability against local ranking variance.

\section{Experimental Details}
\label{exp-details}
\subsection{Hardware and Software Environment}
All experiments were performed on a node equipped with eight NVIDIA RTX~A6000 GPUs~(48\,GB each), running CUDA~12.8 with driver version~570.86.16. The software environment was set up with Python~3.10, PyTorch~2.2.1, and the HuggingFace Transformers library to ensure reproducibility.

\subsection{Training Settings}

All fine-tuning experiments were conducted using parameter-efficient LoRA training~\citep{hu2022lora} on Llama3-8B and Gemma-3-4B, and Mistral-7B backbones. We adopted the PEFT framework with a low-rank adaptation configuration of rank~64, scaling factor~128, and dropout~0.05. Each model was fine-tuned under mixed precision (\texttt{bfloat16}) using the HuggingFace \texttt{Transformers} and \texttt{Accelerate} libraries with \texttt{DeepSpeed ZeRO Stage~2} for distributed optimization. To improve memory efficiency, we enabled \texttt{FlashAttention-2} and non-reentrant gradient checkpointing throughout training. 

We selected the final checkpoint based on the lowest validation loss, which was evaluated every 25 training steps.

\newcommand{\tablescale}{0.88}

\begin{table}[htbp]
\small
\centering
\setlength{\tabcolsep}{10pt}
\renewcommand{\arraystretch}{1.1}
\caption{Key hyperparameters used in all fine-tuning experiments.}
\label{tab:train-hparams}
\scalebox{\tablescale}{%
\begin{tabular}{l l}
\toprule
\textbf{Category} & \textbf{Setting / Value} \\
\midrule
\textit{Dataset} &
\begin{tabular}[t]{@{}l@{}}
Training samples: 20{,}000 \\
Validation samples: 500 \\
Test samples: 500
\end{tabular} \\[4pt]
\midrule
\textit{Optimization} &
\begin{tabular}[t]{@{}l@{}}
Optimizer: AdamW \\
Weight decay: 0.01\\
Learning rate: $5\times10^{-5}$ \\
Scheduler: Cosine (10\% warmup)
\end{tabular} \\[4pt]
\midrule
\textit{Training configuration} &
\begin{tabular}[t]{@{}l@{}}
Epochs: 3 \\
Batch size / device: 4 \\
Gradient accumulation: 4 \\
Max gradient norm: 1.0
\end{tabular} \\
\bottomrule
\end{tabular}%
}
\end{table}

\subsection{Parameter Setting}
The detailed hyperparameter configurations for fine-tuning experiments are summarized in Table~\ref{tab:train-hparams}.

\section{Additional Results}
\label{sec:additional-results}

In this section, we further evaluate the two group based curriculum variants, \textbf{Group Forward Curriculum (GFC)} and \textbf{Group Reverse Curriculum (GRC)}, across all five metric dimensions introduced in Section~\ref{sec:metric-definition}: \textit{Problem Difficulty}, \textit{Model Surprisal}, \textit{Confidence Margin}, \textit{Predictive Uncertainty}, and \textit{Decision Variability}. For each metric, we construct both GFC and GRC schedules and assess their impact on model reasoning performance under identical training settings. The following tables summarize the quantitative results for each metric dimension, providing a detailed comparison between the forward and reverse group based curricula.

\subsection{Problem Difficulty}

\begin{table}[htbp]
\centering
\scriptsize
\setlength{\tabcolsep}{1.5pt}
\renewcommand{\arraystretch}{0.8}

\resizebox{\columnwidth}{!}{%
\begin{tabular}{@{} C{0.12\columnwidth}| C{0.12\columnwidth} | C{0.12\columnwidth}| *{4}{C{0.12\columnwidth}} @{}}
\toprule
\textbf{Model} & \textbf{Strategy} & \textbf{MMQA} & \textbf{ASDiv} & \textbf{GSM8K} & \textbf{MBench} & \textbf{MATH} \\
\midrule

\multirow{9}{*}{\rotatebox[origin=c]{90}{\textbf{Gemma-4B}}}
 & Baseline         & 51.17 & 73.44 & 53.52 & 35.35 & 21.68 \\
 \cmidrule(l{-0pt}r{-0pt}){2-7}
 & RS$_{\uparrow}$   & 56.05 & 76.56 & 55.86 & 33.98 & 24.22 \\
 & RS$_{\downarrow}$ & 52.73 & 75.98 & 53.32 & 33.20 & 22.27 \\
 & SC$_{\uparrow}$   & 56.64 & 76.95 & 55.27 & 30.08 & 26.76 \\
 & SC$_{\downarrow}$ & 53.52  & 73.05 & 52.15 & 36.33 & 22.27 \\
 & CD$_{\uparrow}$   & 52.34 & 77.15 & 53.52 & 35.74 & 23.44 \\
 & CD$_{\downarrow}$ & 52.73 & 75.98 & 50.98 & 35.94 & 22.46 \\
 & ACC$_{\uparrow}$  & 52.34 & 77.15 & 55.47 & 37.50 & 24.61 \\
 & ACC$_{\downarrow}$& 56.04 & 77.93 & 57.81 & 35.74 & 22.66 \\
\midrule

\multirow{9}{*}{\rotatebox[origin=c]{90}{\textbf{Mistral-7B}}}
 & Baseline          & 54.30 & 68.55 & 60.35 & 36.72 & 18.75 \\
 \cmidrule(l{-0pt}r{-0pt}){2-7}
 & RS$_{\uparrow}$   & 54.49 & 71.68 & 60.16 & 35.94 & 16.41 \\
 & RS$_{\downarrow}$ & 59.57 & 72.46 & 63.28 & 36.52 & 16.99 \\
 & SC$_{\uparrow}$   & 53.71 & 72.27 & 60.74 & 36.72 & 16.99 \\
 & SC$_{\downarrow}$ & 53.71 & 74.80 & 61.91 & 37.89 & 18.16 \\
 & CD$_{\uparrow}$   & 54.49 & 70.90 & 57.23 & 39.06 & 16.21 \\
 & CD$_{\downarrow}$ & 53.91 & 72.66 & 62.30 & 38.48 & 18.55 \\
 & ACC$_{\uparrow}$  & 55.47 & 74.02 & 62.50 & 37.70 & 15.04 \\
 & ACC$_{\downarrow}$& 51.95 & 73.24 & 60.74 & 38.28 & 18.16 \\
\midrule

\multirow{9}{*}{\rotatebox[origin=c]{90}{\textbf{Llama-8B}}}
 & Baseline         & 59.57 & 74.02 & 62.50 & 38.09 & 18.55 \\
\cmidrule(l{-0pt}r{-0pt}){2-7}
 & RS$_{\uparrow}$   & 60.35 & 75.00 & 64.26 & 35.35 & 19.73  \\
 & RS$_{\downarrow}$ & 61.52 & 77.54 & 69.92 & 38.67 & 20.70 \\
 & SC$_{\uparrow}$   & 60.16 & 72.07 & 67.97 & 35.74 & 19.53 \\
 & SC$_{\downarrow}$ & 51.95 & 70.51 & 58.40 & 36.52 & 22.85 \\
 & CD$_{\uparrow}$   & 59.18 & 78.32 & 66.60 & 38.28 & 22.27 \\
 & CD$_{\downarrow}$ & 56.84 & 78.52 & 64.26 & 38.87 & 21.29 \\
 & ACC$_{\uparrow}$  & 54.10 & 74.80 & 61.52 & 35.35 & 18.95 \\
 & ACC$_{\downarrow}$& 59.98 & 77.93 & 69.73 & 41.80 & 20.90 \\
\bottomrule
\end{tabular}%
}
\caption{Accuracy (\%) on five mathematical reasoning benchmarks for curricula constructed from \textit{problem difficulty metrics} \textit{(RS, SC, CD, ACC)}. The \textbf{baseline} uses randomly shuffled data. $\uparrow$ and $\downarrow$ indicate \textbf{group-level ascending (GFC)} and \textbf{group-level descending (GRC)} curricula, respectively. See Sec.~\ref{sec:metric-definition} for metric~definitions.}
\label{tab:difficulty-cl-group}
\end{table}

We consider four difficulty metrics for problems: \textit{Reasoning Steps (RS)}, \textit{Symbol Complexity (SC)}, \textit{Comprehension Difficulty (CD)}, and \textit{Acc@\texorpdfstring{$K$}{K} (ACC)}, as described in Section~\ref{sec:p_metric}. The results of Forward CL (FCL) and Reverse CL (RCL) based on these metrics are shown in Table~\ref{tab-2:perplexity-cl}, while their group-based counterparts, Group Forward Curriculum (GFC) and Group Reverse Curriculum (GRC), are reported in Table~\ref{tab:difficulty-cl-group}.

\subsection{Model Surprisal}

Model surprisal is evaluated through \textit{Token-Level Perplexity (TLP)} and \textit{Sequence-Level Perplexity (SLP)}, with formal definitions provided in Section~\ref{model-metric}. FCL and RCL results appear in Table~\ref{tab-2:perplexity-cl}, with their group-based versions (GFC and GRC) in Table~\ref{tab:ppl-cl-group}.

\begin{table}[htbp]
\centering
\scriptsize
\setlength{\tabcolsep}{1.5pt}
\renewcommand{\arraystretch}{0.9}

\resizebox{\columnwidth}{!}{%
\begin{tabular}{@{} C{0.12\columnwidth}| C{0.12\columnwidth}| C{0.12\columnwidth}| *{4}{C{0.12\columnwidth}} @{}}
\toprule
\textbf{Model} & \textbf{Strategy} & \textbf{MMQA} & \textbf{ASDiv} & \textbf{GSM8K} & \textbf{MBench} & \textbf{MATH} \\
\midrule

\multirow{5}{*}{\rotatebox[origin=c]{90}{\textbf{Gemma-4B}}}
 & Baseline         & 51.17 & 73.44 & 53.52 & 35.35 & 21.68 \\
 \cmidrule(l{-0pt}r{-0pt}){2-7}
 & SLP$_{\uparrow}$   & 56.64 & 76.95 & 53.12 & 36.91 & 23.44 \\
 & SLP$_{\downarrow}$ & 51.17 & 74.80 & 52.73 & 38.48 & 21.48 \\
 & TLP$_{\uparrow}$   & 54.30 & 75.00 & 58.98 & 32.62 & 21.68 \\
 & TLP$_{\downarrow}$ & 50.20 & 76.37 & 55.47 & 31.25 & 22.07 \\
\midrule

\multirow{5}{*}{\rotatebox[origin=c]{90}{\textbf{Mistral-7B}}}
 & Baseline         & 54.30 & 68.55 & 60.35 & 36.72 & 18.75 \\
 \cmidrule(l{-0pt}r{-0pt}){2-7}
 & SLP$_{\uparrow}$   & 53.71 & 64.06 & 52.54 & 38.87 & 15.23 \\
 & SLP$_{\downarrow}$ & 54.49 & 75.58 & 63.67 & 38.28 & 15.82 \\
 & TLP$_{\uparrow}$   & 54.30 & 74.80 & 58.20 & 39.06 & 16.21 \\
 & TLP$_{\downarrow}$ & 54.49 & 75.78 & 63.67 & 33.59 & 15.43 \\
\midrule

\multirow{5}{*}{\rotatebox[origin=c]{90}{\textbf{Llama-8B}}}
 & Baseline         & 59.57 & 74.02 & 62.50 & 38.09 & 18.55 \\
 \cmidrule(l{-0pt}r{-0pt}){2-7}
 & SLP$_{\uparrow}$   & 62.50 & 75.78 & 68.16 & 37.89 & 20.31 \\
 & SLP$_{\downarrow}$ & 55.47 & 75.78 & 60.74 & 35.94 & 19.73 \\
 & TLP$_{\uparrow}$   & 56.64 & 74.41 & 59.18 & 37.50 & 19.34 \\
 & TLP$_{\downarrow}$ & 57.42 & 75.59 & 60.94 & 38.28 & 20.90 \\
\bottomrule
\end{tabular}%
}
\caption{Accuracy (\%) on five mathematical reasoning benchmarks under curricula constructed from \textbf{model surprisal metrics} (SLP, TLP). \textbf{Baseline}, $\uparrow$, and $\downarrow$ follow the same notation as in Table~\ref{tab:difficulty-cl-group}.}
\label{tab:ppl-cl-group}
\end{table}

\begin{table}[htbp]
\centering
\scriptsize
\setlength{\tabcolsep}{1.5pt}
\renewcommand{\arraystretch}{1.2}

\begin{tabular}{@{} C{0.12\columnwidth}| C{0.12\columnwidth}| C{0.12\columnwidth}| *{4}{C{0.12\columnwidth}} @{}}
\toprule
\textbf{Model} & \textbf{Strategy} & \textbf{MMQA} & \textbf{ASDiv} & \textbf{GSM8K} & \textbf{MBench} & \textbf{MATH} \\
\midrule

\multirow{3}{*}{\rotatebox[origin=c]{90}{\textbf{Gemma-4B}}}
 & Baseline         & 51.17 & 73.44 & 53.52 & 35.35 & 21.68 \\
 \cmidrule(l{-0pt}r{-0pt}){2-7}
 & LG$_{\uparrow}$   & 52.93 & 73.05 & 54.40 & 34.48 & 21.68 \\
 & LG$_{\downarrow}$ & 56.05 & 75.20 & 53.32 & 34.96 & 22.26 \\
\midrule

\multirow{3}{*}{\rotatebox[origin=c]{90}{\textbf{Mistral-7B}}}
 & Baseline         & 54.30 & 68.55 & 60.35 & 36.72 & 18.75 \\
 \cmidrule(l{-0pt}r{-0pt}){2-7}
 & LG$_{\uparrow}$   & 53.52 & 74.61 & 62.11 & 41.60 & 18.36 \\
 & LG$_{\downarrow}$ & 53.91 & 73.63 & 60.16 & 41.02 & 18.95 \\
\midrule

\multirow{3}{*}{\rotatebox[origin=c]{90}{\textbf{Llama-8B}}}
 & Baseline         & 59.57 & 74.02 & 62.50 & 38.09 & 18.55 \\
 \cmidrule(l{-0pt}r{-0pt}){2-7}
 & LG$_{\uparrow}$   & 61.52 & 75.98 & 64.84 & 39.45 & 21.09 \\
 & LG$_{\downarrow}$ & 58.40 & 73.05 & 66.41 & 38.67 & 21.47 \\
\bottomrule

\end{tabular}

\caption{Accuracy (\%) on five mathematical reasoning benchmarks under curricula constructed from \textbf{confidence margin} (LG). \textbf{Baseline}, $\uparrow$, and $\downarrow$ follow the same notation as in Table~\ref{tab:difficulty-cl-group}.}
\label{tab:lg_cl_group}
\end{table}

\subsection{Confidence Margin}

Confidence margin is measured using the \textit{Logit Gap (LG)} metric, formally defined in Section~\ref{model-metric}. Results for FCL and RCL are presented in Table~\ref{tab:lg_cl_evaluation}, and the corresponding group-based variants, GFC and GRC, are shown in Table~\ref{tab:lg_cl_group}.

\subsection{Predictive Uncertainty}

Predictive uncertainty is measured using the \textit{Token-Level Entropy (TLE)} and \textit{Sequential-Level Entropy (SLE)}, formally defined in Section~\ref{model-metric}. Results for FCL and RCL are presented in Table~\ref{tab:es_et_cl}, and the corresponding group-based variants, GFC and GRC, are shown in Table~\ref{tab:es_et_cl_group}.

\begin{table}[htbp]
\centering
\scriptsize
\setlength{\tabcolsep}{2pt}
\renewcommand{\arraystretch}{1.05}

\begin{tabular}{@{} C{0.12\columnwidth}| C{0.12\columnwidth}| C{0.12\columnwidth}| *{4}{C{0.12\columnwidth}} @{}}
\toprule
\textbf{Model} & \textbf{Strategy} & \textbf{MMQA} & \textbf{ASDiv} & \textbf{GSM8K} & \textbf{MBench} & \textbf{MATH} \\
\midrule

\multirow{5}{*}{\rotatebox[origin=c]{90}{\textbf{Gemma-4B}}}
 & Baseline         & 51.17 & 73.44 & 53.52 & 35.35 & 21.68 \\
 \cmidrule(l{-0pt}r{-0pt}){2-7}
 & SLE$_{\uparrow}$   & 54.49 & 76.76 & 57.81 & 33.01 & 26.37 \\
 & SLE$_{\downarrow}$ & 55.27 & 76.37 & 53.21 & 26.56 & 25.98 \\

& TLE$_{\uparrow}$   & 52.54 & 74.02 & 52.34 & 37.30 & 22.07 \\
 & TLE$_{\downarrow}$ & 52.73 & 74.80 & 51.95 & 37.89 & 22.85 \\
 
\midrule
\multirow{5}{*}{\rotatebox[origin=c]{90}{\textbf{Mistral-7B}}}
 & Baseline         & 54.30 & 68.55 & 60.35 & 36.72 & 18.75 \\
 \cmidrule(l{-0pt}r{-0pt}){2-7}
 & SLE$_{\uparrow}$   & 53.91  & 68.55 & 61.91 & 38.67 & 17.77 \\
 & SLE$_{\downarrow}$ & 54.30 & 75.39 & 64.65 & 39.65 & 13.67 \\
 & TLE$_{\uparrow}$   & 51.56 & 71.48 & 56.84 & 39.45 &  15.23\\
 & TLE$_{\downarrow}$ & 49.61 & 72.85 & 59.38 & 39.26 & 14.45 \\
\midrule

\multirow{5}{*}{\rotatebox[origin=c]{90}{\textbf{Llama-8B}}}
 & Baseline         & 59.57 & 74.02 & 62.50 & 38.09 & 18.55 \\
 \cmidrule(l{-0pt}r{-0pt}){2-7}
 & SLE$_{\uparrow}$   & 58.20 & 78.32 & 66.02 & 38.09 & 20.12 \\
 & SLE$_{\downarrow}$ & 53.71 & 71.68 & 61.13 & 38.67 & 19.34 \\
 & TLE$_{\uparrow}$   & 55.66 & 77.54 & 63.48 & 39.06 & 19.53 \\
 & TLE$_{\downarrow}$ & 54.69 & 74.22 & 62.50 & 39.45 & 22.66 \\
\bottomrule
\end{tabular}

\caption{Accuracy (\%) on five mathematical reasoning benchmarks under curricula constructed from \textbf{predictive uncertainty metrics} (SLE, TLE). \textbf{Baseline}, $\uparrow$, and $\downarrow$ follow the same notation as in Table~\ref{tab:difficulty-cl-group}.}
\label{tab:es_et_cl_group}
\end{table}

\subsection{Decision Variability}

Decision variability is evaluated using the \textit{Acc@\texorpdfstring{$K$}{K}} metric (VACC), as defined in Section~\ref{model-metric}. Results for FCL and RCL are reported in Table~\ref{tab:vacc_cl}, while their group-based variants, GFC and GRC, are presented in Table~\ref{tab:vacc_cl_group}.

\begin{table}[htbp]
\centering
\scriptsize
\setlength{\tabcolsep}{2pt}
\renewcommand{\arraystretch}{1.2}

\begin{tabular}{@{} C{0.12\columnwidth}| C{0.12\columnwidth}| C{0.12\columnwidth}| *{4}{C{0.12\columnwidth}} @{}}
\toprule
\textbf{Model} & \textbf{Strategy} & \textbf{MMQA} & \textbf{ASDiv} & \textbf{GSM8K} & \textbf{MBench} & \textbf{MATH} \\
\midrule

\multirow{3}{*}{\rotatebox[origin=c]{90}{\textbf{Gemma-4B}}}
 & Baseline         & 51.17 & 73.44 & 53.52 & 35.35 & 21.68 \\
 \cmidrule(l{-0pt}r{-0pt}){2-7}
 & VACC$_{\uparrow}$   & 48.44 & 75.39 & 49.41 & 33.20 & 23.63 \\
 & VACC$_{\downarrow}$ & 56.25 & 71.88 & 57.03 & 25.78 & 18.75 \\
\midrule

\multirow{3}{*}{\rotatebox[origin=c]{90}{\textbf{Mistral-7B}}}
 & Baseline         & 54.30 & 68.55 & 60.35 & 36.72 & 18.75 \\
 \cmidrule(l{-0pt}r{-0pt}){2-7}
 & VACC$_{\uparrow}$   & 54.49 & 68.55 & 60.35 & 40.43 & 15.82 \\
 & VACC$_{\downarrow}$ & 52.93 & 73.44 & 61.52 & 43.16 & 16.99 \\
\midrule

\multirow{3}{*}{\rotatebox[origin=c]{90}{\textbf{Llama-8B}}}
 & Baseline         & 59.57 & 74.02 & 62.50 & 38.09 & 18.55 \\
 \cmidrule(l{-0pt}r{-0pt}){2-7}
 & VACC$_{\uparrow}$   & 49.61 & 76.37 & 58.01 & 32.03 & 49.41 \\
 & VACC$_{\downarrow}$ & 58.20 & 78.32 & 65.23 & 40.43 & 58.20 \\
\bottomrule
\end{tabular}

\caption{Accuracy (\%) on five mathematical reasoning benchmarks under curricula constructed from \textbf{decision variability} (VACC). \textbf{Baseline}, $\uparrow$, and $\downarrow$ follow the same notation as in Table~\ref{tab:difficulty-cl-group}.}
\label{tab:vacc_cl_group}
\end{table}

\subsection{Result Analysis of FCL/RCL and GFC/GRC}

We compare fine-grained curricula (FCL/RCL) with their group-based variants (GFC/GRC). Overall, GFC and GRC yield more stable and consistent reasoning performance across benchmarks and model scales.

\paragraph{(a) Stability and direction.} 
GFC/GRC produce smoother results and fewer fluctuations between curriculum directions. Grouping samples before scheduling reduces local ranking noise, improving directional consistency—e.g., GFC benefits the \textit{RS} metric on \textit{Gemma-4B}, while GRC enhances \textit{ACC} on \textit{Llama-8B}.

\paragraph{(b) Robustness of effects.} 
While FCL/RCL occasionally reach higher peaks, their performance varies more with direction choice. GFC/GRC achieve more balanced gains, suggesting that grouped scheduling stabilizes optimization and mitigates overfitting to specific ranking orders.

\paragraph{(c) Model-scale sensitivity.} 
Smaller models (\textit{Gemma-4B}, \textit{Mistral-7B}) show larger differences between FCL/RCL and GFC/GRC, whereas larger ones (\textit{Llama-8B}) are less sensitive to ranking granularity. This indicates that coarser grouping becomes increasingly effective with model size.

\noindent\textbf{Summary.} 
Group-based curricula improve the stability, robustness, and interpretability of curriculum learning by suppressing noisy ranking signals and yielding more consistent training behavior.

\end{document}